\documentclass[sigconf]{acmart}

\usepackage{multirow}
\usepackage{booktabs} 
\usepackage{listings}
\usepackage[acronym]{glossaries}
\usepackage{changepage}
\usepackage[T1]{fontenc}
\usepackage[utf8]{inputenc}
\usepackage[english]{babel}
\usepackage{vhistory}
\usepackage{graphicx}
\usepackage{tabularx}
\usepackage{siunitx}
\usepackage{xcolor,colortbl}
\usepackage[linesnumbered,ruled,vlined]{algorithm2e}

\copyrightyear{2020}
\acmYear{2020}
\setcopyright{acmcopyright}
\acmConference[GECCO '20]{Genetic and Evolutionary Computation Conference}{July 8--12, 2020}{Cancún, Mexico} 
\acmBooktitle{Genetic and Evolutionary Computation Conference (GECCO '20), July 8--12, 2020, Cancún, Mexico}
\acmPrice{15.00}
\acmDOI{10.1145/3377930.3390158}
\acmISBN{978-1-4503-7128-5/20/07}

\begin{document}
\title{AutoLR: An Evolutionary Approach to Learning Rate Policies}

\author{Pedro Carvalho }
\orcid{xxxx}
\affiliation{%
	\institution{University of Coimbra, CISUC, DEI}
	\streetaddress{DEI, Polo 2, Pinhal de Marrocos}
	\city{Coimbra} 
	\state{Portugal} 
	\postcode{3030-290}
}
\email{pfcarvalho@student.dei.uc.pt}

\author{Nuno Louren\c{c}o}
\orcid{0000-0002-2154-0642}
\affiliation{%
	\institution{University of Coimbra, CISUC, DEI}
	\streetaddress{DEI, Polo 2, Pinhal de Marrocos}
	\city{Coimbra} 
	\state{Portugal} 
	\postcode{3030-290}
}
\email{naml@dei.uc.pt}

\author{Filipe Assun\c{c}\~{a}o}
\orcid{0000-0002-0915-8475}
\affiliation{%
	\institution{University of Coimbra, CISUC, DEI}
	\institution{University of Lisbon}
	\streetaddress{DEI, Polo 2, Pinhal de Marrocos}
	\city{Coimbra} 
	\state{Portugal} 
	\postcode{3030-290}
}
\email{fga@dei.uc.pt}

\author{Penousal Machado}
\orcid{0000-0002-6308-6484}
\affiliation{%
	\institution{University of Coimbra, CISUC, DEI}
	\streetaddress{DEI, Polo 2, Pinhal de Marrocos}
	\city{Coimbra} 
	\state{Portugal} 
	\postcode{3030-290}
}
\email{machado@dei.uc.pt}


\begin{abstract}
The choice of a proper learning rate is paramount for good Artificial Neural Network training and performance. In the past, one had to rely on experience and trial-and-error to find an adequate learning rate. Presently, a plethora of state of the art automatic methods exist that make the search for a good learning rate easier. While these techniques are effective and have yielded good results over the years, they are general solutions. This means the optimization of learning rate for specific network topologies remains largely unexplored. This work presents AutoLR, a framework that evolves Learning Rate Schedulers for a specific Neural Network Architecture using Structured Grammatical Evolution. The system was used to evolve learning rate policies that were compared with a commonly used baseline value for learning rate. Results show that training performed using certain evolved policies is more efficient than the established baseline and suggest that this approach is a viable means of improving a neural network's performance. 
\end{abstract}

%
%



\maketitle

\newglossaryentry{ANN}{
    name=ANN,
    description={Artificial Neural Network}
}
\newglossaryentry{AI}{
    name=AI,
    description={Artificial Intelligence}
}
\newglossaryentry{LR}{
    name=LR,
    description={Learning Rate}
}
\newglossaryentry{EA}{
    name=EA,
    description={Evolutionary Algorithm}
}
\newglossaryentry{GE}{
    name=GE,
    description={Grammatical Evolution}
}
\newglossaryentry{SGE}{
    name=SGE,
    description={Structured Grammatical Evolution}
}
\newglossaryentry{DNN}{
    name=DNN,
    description={Deep Neural Network}
}
\newglossaryentry{CNN}{
    name=CNN,
    description={Convolutional Neural Network}
}
\newglossaryentry{DENSER}{
    name=DENSER,
    description={Deep Evolutionary Network Structured Representation}
}
\section{Introduction}
\label{sec:introduction}
The study of Artificial Neural Networks (\glspl{ANN}) is a field in modern Artificial Intelligence (\gls{AI}). These networks' defining characteristic is that they are able to learn how to perform a certain task when provided with an appropriate architecture, data and resources. The networks have a set of internal parameters known as \textbf{weights} and \textbf{training} is the process through which they are modified so that the network is able to solve a given problem. Fine-tuning the weights of \glspl{ANN} is crucial in order to obtain a consistently useful system. There are several parameters that regulate training, one the most important parameters is the \textbf{learning rate}. In fact, and according to \cite[p.~424]{goodfellow2016deep}, if we only have the chance to modify one hyperparameter, the focus should be on the learning rate.


The learning rate determines the magnitude of the changes that are made to the weights. Consequently, the choice of an adequate learning rate is paramount for effective training. When the value of the learning rate is too small the network will be unable to make impactful changes to its weights, making the training slow. On the other hand, if the learning rate is too high the system will make radical changes even in response to small mistakes, causing inconsistent and unpredictable behaviour. On top of this, research suggests that the best training results are achieved by adjusting the learning rate over the course of the training process~\cite{Senior2013}. One way to make these adjustments during training is by updating the learning rate as training progresses. The functions responsible for these adjustments are known as \textbf{learning rate policies}. There is subset of these functions known as learning rate schedulers, i.e., functions that are periodically called during training and return a new learning rate based on multiple training characteristics, such as the current learning rate or the number of performed iterations.

The main objective of this work is to devise an approach that is able to evolve learning rate policies for specific neural network architectures, in order to improve its performance. In concrete, we developed AutoLR, a system that allows us to study the viability of this approach and how it may contribute to the field of learning rate optimization as a whole.
Learning rate policies can take many different shapes \cite{smith2017cyclical}, and therefore it will be notable if our system is capable of automatically discovering functions that are variations of the ones found in the literature. Such a result is interesting because if this approach is able to evolve solutions that are widely accepted it is possible that these same ideas can be used to find still undiscovered, better methods. 
We are also interested in inspecting the evolved schedulers, and comparing them with human-designed schedulers to obtain meaningful insights. The contributions of this paper are:
\begin{itemize}
    \item The proposal of AutoLR, a framework based on SGE that performs automatic optimisation for learning rate schedulers.
    \item The design, test and analysis of experiments that validate the use evolutionary algorithms to optimise learning rate schedulers.
    \item We show that the evolved policies are competitive and have characteristics that allow then to thrive in the problems at hand.
\end{itemize}

The remainder of the paper is organised as follows. Section~\ref{sec:related_work} introduces the background concepts and surveys the key-works related to the optimisation of learning rate schedulers. Section~\ref{sec:AutoLR} describes AutoLR, the methodology proposed for the evolution of learning rate schedulers. Section~\ref{sec:experimentation} details the experimental setup and discusses the experimental results. Finally, Section~\ref{sec:conclusions} summarises the main conclusions of the paper and addresses future work.

\section{Related Work}
\label{sec:related_work}

This section provides the necessary context for the reader to understand the rest of the paper. Section~\ref{sec:ann} introduces Artificial Neural Networks; Section~\ref{sec:sge} details Structured Grammatical Evolution; and Section~\ref{sec:lr_optimization} surveys works related to learning rate optimization and learning rate schedulers.

\subsection{Artificial Neural Networks}
\label{sec:ann}

\begin{figure}[t!]
    \centering 
    \includegraphics[width=0.25\textwidth,keepaspectratio]{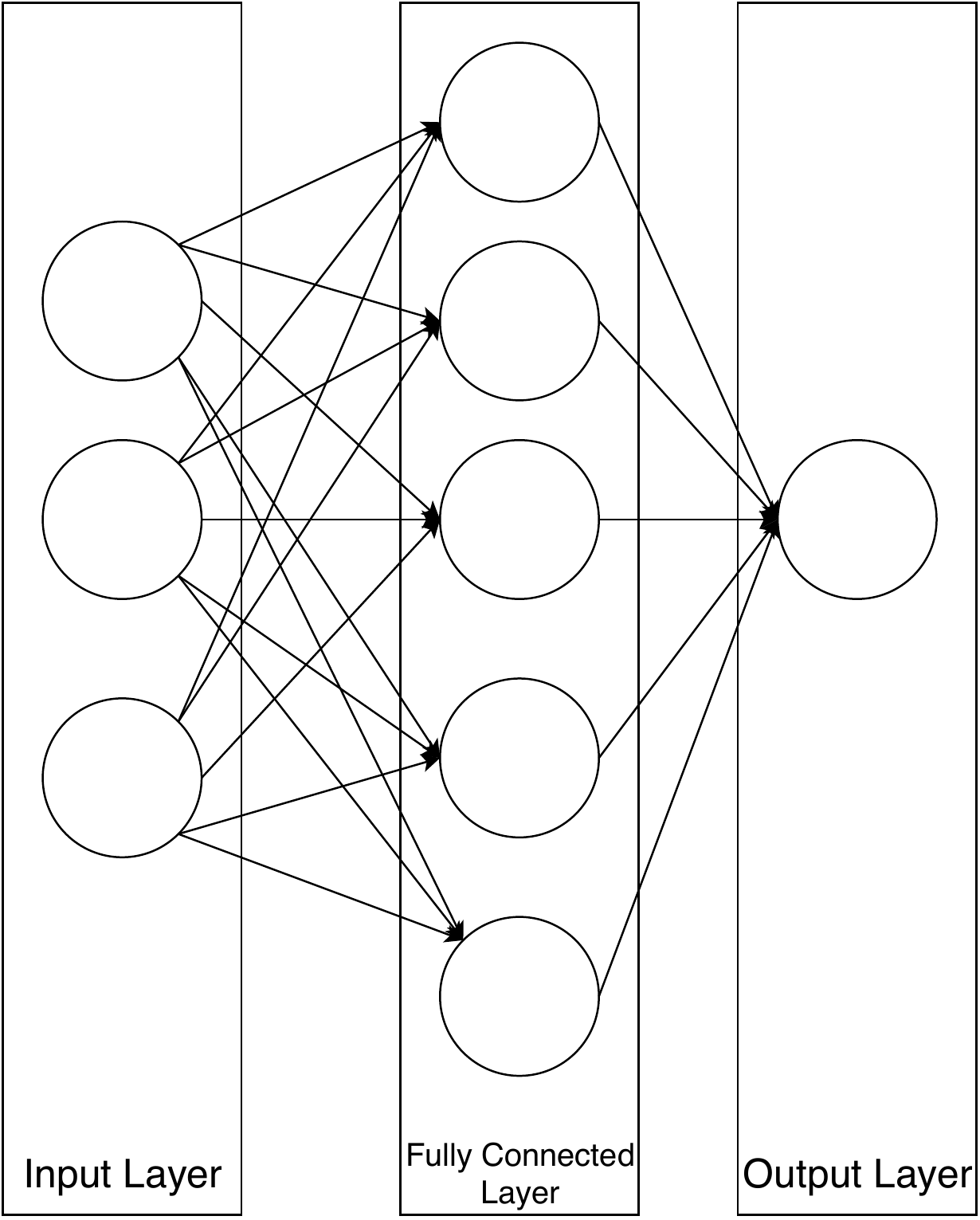}
    \caption{Example of an Artificial Neural Network.}
    \label{fig:example_neural_network} 
\end{figure}

Artificial Neural Networks (\glspl{ANN}) are a machine learning approach that draws inspiration  in the biological neural networks seen in nature in order to create a computing system that is able to learn. These systems are comprised by a set of nodes (known as neurons) and edges (known as synaptic weights). An example of the general structure of an ANN is depicted in Figure~\ref{fig:example_neural_network}.

Although these networks can have different architectures (e.g., LSTM \cite{Hochreiter1997}, ResNet \cite{He2016}) we will, without loss of generality, focus on feed-forward \glspl{ANN}. In these models the nodes are grouped into separate layers connected sequentially. These layers are flanked by an input and output layer which are responsible for receiving the data that the network will process, and yield the result of the network's calculations, respectively. Edges are directional connections between two nodes from different layers and are the means through which information travels through the network. Each node performs an operation (i.e., a mathematical function) on the values it receives from the previous layer and sends the new value to all nodes it is connected to in the next layer. Every edge has a weight that scales the value it carries, i.e., the value that a node outputs is always adjusted before it is provided to the nodes in the next layer. 

These networks can be used to solve tasks of many different types. The ideas that will be presented in this work can be widely applied to different types of \glspl{ANN}. Without loss of generality we will focus in the optimisation of a learning rate scheduler for a supervised learning classification problem. In supervised learning the system is tasked with learning a function that can separate data instances into their respective classes. To achieve this the network is provided with a set of labeled instances.

The training of \glspl{ANN} is an iterative process where the network compares its attempted classifications of a subset of examples with the expected ones and adjusts its weights to get closer to the correct results. There is a function -- known as loss function -- that compares the classification and measures how incorrect the network's output was. The size of the changes made to the weights is partially given by the error returned by the loss function (a larger error leads to larger changes). Another parameter, the \textbf{Learning Rate} (learning rate), determines the magnitude of the adjustments that are made to the weights. The learning rate is the main subject of this paper. For more details on \glspl{ANN} refer to~\cite{10.5555/2810085}.

\begin{figure}[t!]
    \centering 
    \includegraphics[width=0.5\textwidth,keepaspectratio]{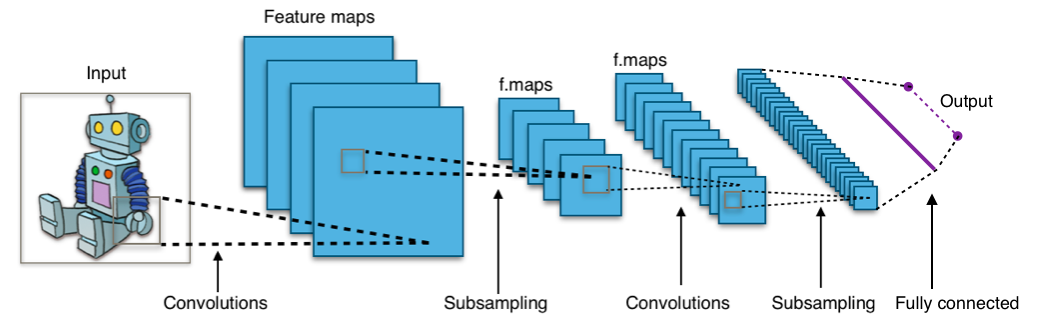}
    \caption{Example of the architecture of a Convolutional Neural Network.} 
    \label{fig:ex_cnn} 
\end{figure}

Deep Neural Networks (\glspl{DNN}) are a subset of \glspl{ANN} notable for being able to perform representation learning, and consequently the networks are able to automatically extract the features required to solve the problem. This is often associated to the need for deeper architectures, i.e., a greater number of hidden-layers. This allows the networks to possibly solve solve harder problems. In the current work we focus on Convolutional Neural Networks (\glspl{CNN})~\cite{Fukushima1980}, a \gls{DNN} topology that is known to work well on spatially-related data (e.g., image). An example of the architecture of \glspl{CNN} is shown in  Figure~\ref{fig:ex_cnn}. Two layer types are commonly used in \glspl{CNN}: convolutional and pooling layers. More details can be found in~\cite{lecun1998gradient}


\subsection{Structured Grammatical Evolution}
\label{sec:sge}

In the current work we will perform the optimisation of learning rate schedulers using \gls{SGE}. \gls{SGE} is a variant of \gls{GE}~\cite{ONeill2001} that uses an altered genotype representation to address the main limitations of \gls{GE}: low locality and high redundancy. In \gls{GE} the genotype is encoded as a single list of integers, where each integer encodes a grammatical expansion possibility. Contrary, in \gls{SGE} there is a separate list for each non-terminal symbol; this avoids the need for the modulo operation when performing the genotype to phenotype mapping.

These approaches add another layer of decision making however, namely in the form of the grammar design \cite{DBLP:conf/eurogp/KeijzerORC02}. The grammar used for any \gls{GE} experiment will define what kind of programs the engine is able to create and this has many implicit consequences. The most obvious one is that the provided grammar must encompass solutions that can solve the problem at hand. While this seems trivial it must be understood that not knowing the composition of the desired program is one of the main motivations to use this type of system in the first place. This also means, however, that the grammar specificity can be increased as more knowledge of the problem is available, aiding the search process. 

This specific type of \gls{EA} is suited for this work because the functions we are looking to evolve are very specific. This means that our domain knowledge is high, and there is a strong understanding of what our desired program is like. As previously mentioned we can use this knowledge to create a grammar that enhances results by narrowing the search space. 
An in-depth explanation of these algorithms can be found in \cite{10.5555/2810085, ryan2018handbook}.

\subsection{Learning Rate Optimization}
\label{sec:lr_optimization}

In the context of this work, \textbf{hyper-parameters} are the set of parameters that configure an \gls{ANN} and its training strategy. The \textbf{learning rate} is one of such parameters and its role is to scale the changes made to the network weights during training. Research suggests that hyper-parameter optimization is effective in improving the system's performance without adding complexity \cite{Bergstra2011}. 

\subsubsection{Static Learning Rate}

The traditional approach is to use a single learning rate for the entire training process \cite{reed1999neural}. Under these circumstances all optimization must be done before training starts. Oftentimes the programmer must rely on expertise and intuition in order to \textit{guess} adequate learning rate values. While automatic solutions to this problem exist they are, to the best of our knowledge, either comparable to manual optimization \cite{Bergstra2011} or non-trivial in implementation \cite{Bergstra2012}. Much of the difficulty of finding a convenient solution to this issue stems from the fact that hyper-parameters are inter-dependent \cite{Breuel2015}. This means that even when an ideal learning rate is found there is no guarantee that this value remains optimal (or even usable) as the other parameters are tweaked.

\subsubsection{Dynamic Learning Rate}

The reasons stated in the previous section make the use of a static learning rate a possible drawback. It is desirable that the method we are using to determine our learning rate is robust enough that performance does not dip with every change to the system. In order to increase flexibility we would ideally have a method to change the learning rate as training progresses, i.e., even if the initial value is not adequate the system has a chance to correct its course. This strategy will be referred to as a \textbf{dynamic learning rate}. The most uncomplicated policy for varying the learning rate can be inferred intuitively. It is expected that as training progresses the \gls{ANN}'s performance gradually improves as it gets better at solving the task at hand. If the system is potentially closer to its objective it seems desirable that it does not stray from its course. This is to say that, in order to improve, the network requires progressively finer tuning; this can be achieved with a \textbf{decaying learning rate} (meaning that the learning rate decreases as learning progresses). There are some issues that are frequently encountered during training that make this approach not ideal however. Better performance is rarely an indicator that the network is closer to a perfect solution. 
Using a decaying learning rate leaves the system susceptible to early stagnation in a local optimum. This is not ideal despite the fact that a local optimum is sufficient for most situations as this approach can lead to early stagnation if applied incorrectly. Despite these limiting factors decaying learning rates can lead to improvement over static ones as seen in \cite{Senior2013}.

In order to expand on these ideas we need to apply the concepts of \textbf{exploration} and \textbf{exploitation}. These refer to the two complementary strategies that can be used in heuristic optimization. Exploration is the idea of using a mechanism that helps the algorithm \textit{explore} solutions that do not seem as promising in an attempt to avoid falling into a local optimum. The contrasting technique is exploitation, in this strategy we adjust our approach to make sure the algorithm is able to find the local optimum (once it reaches a promising region). Finding a proper balance between these two strategies is crucial for further improvement of the dynamic learning rate. \textit{Smith et. al.} propose the use of a \textit{cyclic learning rate} in \cite{smith2017cyclical}. Their approach fluctuates the learning rate between a maximum and a minimum bound. While the system uses no information about whether or not it is stuck by periodically increasing the learning rate it is able to explore the search space more effectively. This technique is consequently less vulnerable to early stagnation than decaying learning rate policies. This method is, to the best of our knowledge, the most efficient use of dynamic learning rates.

\subsubsection{Adaptive Learning Rate}

Further improvements in this area can still be achieved if the system responsible for assigning the learning rate has access to information throughout training. This means that we will now study algorithms that can acknowledge when training is stagnating as it is happening. From this point onward we refer to these methods as \textbf{adaptive learning rates}. 

These techniques unlock one more option of optimization. So far we have been working with a single value learning rate but with this extra information it is desirable to use a vector of values instead. Consider the following scenario, an \gls{ANN} is being trained for 100 generations with a single value adaptive learning rate. One specific weight of the network reaches a near optimal value within the first 5 generations, but all of the others are still off the mark. An adaptive learning rate recognizes this and has to decide what is the ideal learning rate value for the next generation. On the one hand, using a small learning rate will benefit the fine tuning of the node that is already performing well. A larger learning rate, on the contrary, will allow the sub optimal weights to find better values. Using vectors of learning rates allows the system to have a learning rate value for each weight, making the most out of these nuanced situations \cite{Jacobs1988}. Several algorithms \cite{Zeiler2012,Duchi2010,Kingma2014} have been built on this theoretical foundation and these systems are the best learning rate policies we know of.

\section{AutoLR: Evolution of Learning Rate Schedulers}
\label{sec:AutoLR}

AutoLR is a framework created to apply evolutionary algorithms to learning rate policy optimization. While \gls{SGE} is used to handle the evolutionary processes, the system's novelty comes from using the algorithm to explore new possibilities in the learning rate policy search space. This is achieved through the design of a grammar that is able to effectively navigate part of this space and a fitness function that can accurately measure each policy's quality.

\subsection{Evolved Policies}
The scope of this work is limited to evolving learning rate schedulers. We define learning rate schedulers as it is done in the Keras\cite{keras} library. Learning rate schedulers are functions that are called periodically during training (each epoch, in this case) and update the learning rate value. In other words, we are evolving the initial learning rate and the ensuing variation function. These functions' inputs are comprised of the learning rate of the previous epoch and the number of performed epochs. This function returns a single learning rate for all dimensions. Using the terminology established so far, this means \textbf{the evolved policies can be either a static or dynamic learning rate solution}. It is important to define the range of our solutions as this establishes what conventional techniques we should be kept in mind during analysis.

\begin{figure}[t!]
    \centering 
    \includegraphics[width=0.4\textwidth,keepaspectratio]{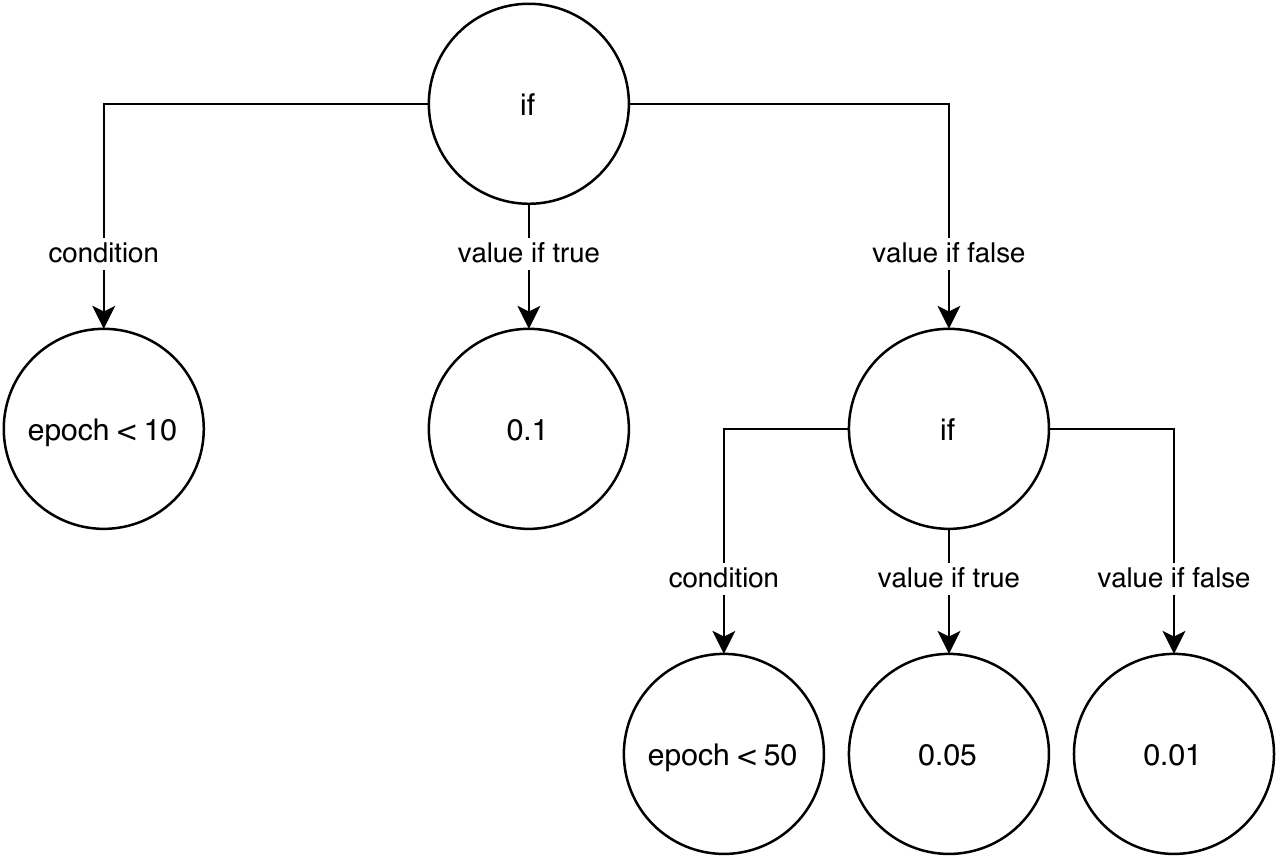}
    \caption{Example of a learning rate scheduler.}
    \label{fig:lr_scheduler} 
\end{figure}

Figure~\ref{fig:lr_scheduler} depicts an example of a learning rate scheduler. In this case the \gls{ANN} will train using a learning rate of 0.1 for the first 10 epochs as this is when the condition \textit{epoch $<$ 10} is met. This learning rate will be used until the 10th epoch is reached, at which point the learning rate scheduler will automatically decrease the learning rate to 0.05. Following the same rationale, after the 50th epoch the learning rate to use is 0.01. The search space that we consider is detail on the next sub-section.

\subsection{Grammar}

\begin{figure}[t!]
    \centering
    \begin{align*}
        {<}\text{expr}{>} ::= & \, \text{if\_func(}{<}\text{logic\_expr}{>}\text{, }{<}\text{expr}{>}\text{, }{<}\text{expr}{>}\text{)} \\
                              & \, | \, {<}\text{lr\_const}{>} \\
        {<}\text{logic\_expr}{>} ::= & \, \text{learning\_rate} \, {<}\text{logic\_op}{>} \, {<}\text{lr\_const}{>} \\
                                     & \, | \, \text{epoch} \, {<}\text{logic\_op}{>} \, {<}\text{ep\_const}{>} \\
        {<}\text{logic\_op}{>} ::= & \, < \, | \, \leq  \, | \, > \, | \, \geq \\ 
        {<}\text{lr\_const}{>} ::= & \, 0.0001 \, | \, 0.00110909 \, | \, 0.00211818 \, | \, 0.00312727 \, |  \\
                                   & \, \ldots \\
                                   & \, 0.09596364 \, | \, 0.09697273 \, | \, 0.09798182 \, | \, 0.09899091 \, |\\ 
                                   & \, 0.1 \\
        {<}\text{ep\_const}{>} ::= & \, 1  \, | \, 2 \, | \, 3 \, | \, 4 \, | \, 5 \, | \, 6 \, | \, 7 \, | \, 8 \, | \, 9 \, | \, 10 \\
                                   & \, \ldots \\
                                   & 91 \, | \,92 \, | \, 93 \, | \, 94 \, | \, 95 \, | \, 96 \, | \, 97 \, | \, 98 \, | \, 99 | \, 100 \\
    \end{align*}
    \caption{Grammar used for the optimisation of learning rate schedulers.}
    \label{fig:cnn_grammar}
\end{figure}

The grammar (Figure~\ref{fig:cnn_grammar}) defines the search space of the learning rate schedulers. The individuals created by this grammar will typically resolve into a sequence of chained if-else conditions (created by the \textit{logic\_expr} production) that once evaluated yield a learning rate (provided by the terminals in \textit{lr\_const}). This means that the system is creating dynamic learning rate policies most of the time. A notable exception to this is that the system can resolve the initial \textit{expr} production into a \textit{lr\_const}, creating a static learning rate policy.

An \textit{if\_func} is a simple function that does the same as a regular if-then-else construct. Since the code for this system was written in Python this function was created so all individuals could be described in a single line that can be read easily by the user. The code for this function is shown in Algorithm~\ref{if_code}.

The conditions used by \textit{if\_func} are generated by \textit{logic\_expr}. This production will compare one of the input variables (learning\_rate, epoch) with the corresponding constants (lr\_const and ep\_const, respectively) using one of several logical operators from \textit{logic\_op}. \textit{logic\_op} includes all logical operators with the exception of equality (==) and inequality (!=). Conditions using these operators are too specific since they only return a different value for a single constant. This means that, in the vast majority of situations, conditions using these operators do not change the policy's behaviour. This makes them unable to contribute meaningfully to the evolutionary process.

The constants chosen for \textit{lr\_const} and \textit{ep\_const} are 100 evenly spaced values between the minimum and maximum value for each of the variables. It should be noted that these production rules have been abridged in the figure. Only a few of the lowest and highest possible values are shown so that the range is accurately portrayed whilst keeping the figure brief. 
Our training starts in epoch 1 and ends in epoch 100, since we are also using 100 values for our constant we used every possible epoch value (every natural number from 1 to 100) for \textit{ep\_const}. \textit{lr\_const} values are more complicated as there is an infinite number of valid learning rates. We keep the values of the learning rate bounded between 0.001 and the 0.1 as all values in this range are suitable for training.

This grammar is capable of creating a large variety of individuals despite its simplicity. While it is not possible for our trees to exactly recreate the dynamic solution functions mentioned in Section~\ref{sec:lr_optimization} they can reproduce approximated versions that exhibit similar behaviour.

\begin{algorithm}
\SetKwInOut{Params}{params}
\Params{condition, state1, state2}

\eIf{condition}{\Return state1\;}{\Return state2\;}
  \caption{Template of the code used to implement the if\_func routine.}
 \label{if_code}
\end{algorithm}

\subsection{Fitness Function}

As the main hypothesis implies we are looking to evolve learning rate policies. This means that we will be using an \gls{EA} on a population of learning rate policies. Additionally, our hypothesis demands that an individual's fitness must be some measure of the network's performance when trained using that specific solution. This is necessary since if the evolutionary process is not successful, its results will not address the question we posed. 

We decided that the best way to assess a policy's performance was through the function seen in Algorithm \ref{fitness_function}. That is, we train the network and assess its performance using the accuracy metric.

\begin{algorithm}
\SetKwInOut{Params}{params}
\Params{network, learning\_rate\_policy, training\_data, test\_data}
trained\_network $\leftarrow{}$ train(network, learning\_rate\_policy, training\_data)\;
fitness\_score $\leftarrow{}$ get\_test
\_accuracy(trained\_network, test\_data)\;
\Return fitness\_score\;

\caption{Simplified version of the fitness function used to evaluate a learning rate policy}
\label{fitness_function}
\end{algorithm}

To elaborate on the algorithm above, our fitness function will use 4 components
\begin{itemize}
    \item \textbf{network} - The \gls{ANN} we are optimizing the learning rate scheduler for. This network is the same throughout the entire evolutionary run.
    \item \textbf{learning\_rate\_policy} - An evolved learning rate scheduler that we want to evaluate. 
    \item \textbf{training/test\_data} - This is the data of the problem the \gls{ANN} will be attempting to solve. As the name implies, training data is used for training. Test data is a separate set of examples that are used to evaluate the network's performance once training is complete. In the actual fitness function the training data is further split into training and validation (see \ref{sec:experimental_setup}) but this distinction will temporarily be omitted for explanation's sake.
\end{itemize}

The evaluation function has two phases. First, the network must be trained, this is where the policy we are evaluating will affect the process. The \textbf{train} function returns the network provided with its weights changed through the training process. We could at this point also retrieve the best performance the network achieved during training. We do not take this approach as it is not the most accurate measure of an \gls{ANN}'s real effectiveness. The objective of the training is that the network learns a set of weights that solve the proposed problem. The data used for training is only a sample of all possible inputs. As training progresses a network becomes gradually too attached to the training data, this is known as \textbf{overfitting}. Overfitting means that the network is to constraint to the training data, and does not represent the general learning problem. This happens since data will often have some noise (i.e. information that is not important to solve the task). It is not desirable for the network to learn to produce solutions based on this noise as that will hurt its performance when dealing with inputs not included in training. Consequently, we measure the effectiveness of training by how well the network performs on a second set of data that it has not come into contact with. We call this second set the test data.

Every policy will be evaluated using the same network and training data meaning that the learning rate scheduler is the only varying component between individuals. Since all other hyper-parameters are fixed, and the used datasets are balanced, we consider the result of evaluating the trained network's \textbf{accuracy} on the test data to be an adequate measure of the policy's fitness. 

In the context of our work, learning rate policies are executable computer code. We will be using the Python language specifically as it has vast support for \gls{ANN} handling through the \textbf{Tensorflow} \cite{tensorflow2015-whitepaper} library. An \gls{EA} is also needed for our system, we chose to use \gls{GE}-based evolutionary engine as it gives us a flexible and readable means of defining the problem space in the form of grammars. In particular, we chose \gls{SGE} \cite{Lourenco2016} for its Python implementation and superior results over regular \gls{GE}. Our hypothesis also demands a mindful choice of network architecture. Since we are looking for optimization in specific scenarios, we want to avoid generic architectures. We therefore decided to use a CNN model evolved specifically for image classification obtained from Deep Evolutionary Network Structured Representation (DENSER)~\cite{assuncao2018gpem}.

\section{Experimentation}
\label{sec:experimentation}

The objective of this work is to promote the automatic optimisation of learning rate schedulers for a fixed-topology network. Section~\ref{sec:network} introduces the topology of the used network; Section~\ref{sec:dataset} details the dataset; Section~\ref{sec:experimental_setup} describes the experimental setup; and Section~\ref{sec:experimental_results} analyses and discusses the experimental results.

\subsection{Network Architecture}
\label{sec:network}
\begin{figure}[t!]
    \centering 
    \includegraphics[width=0.35\textwidth,keepaspectratio]{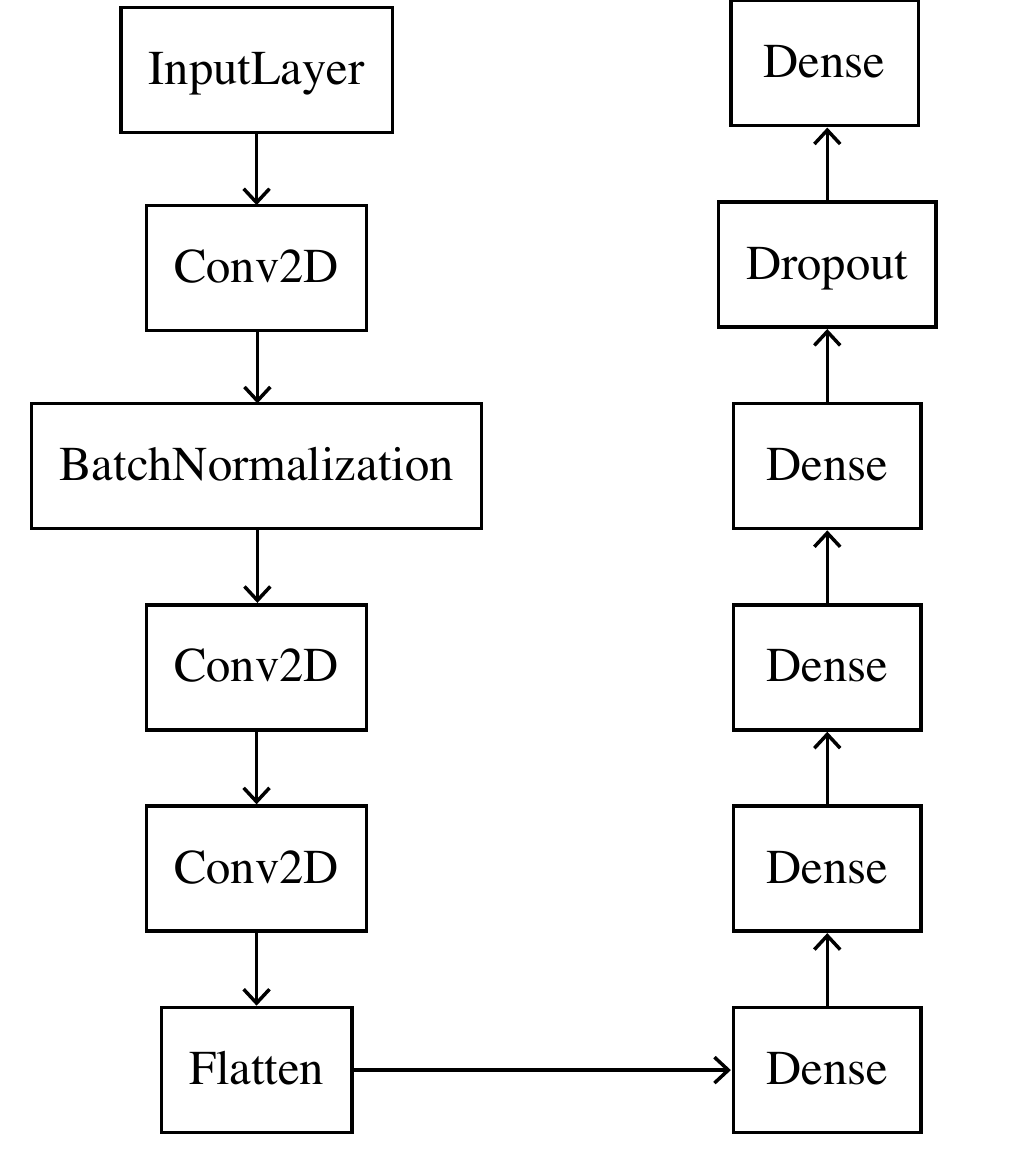}
    \caption{Topology of the used CNN.}
    \label{fig:cnn_topology} 
\end{figure}

The network architecture we used was automatically generated using \gls{DENSER} \cite{assuncao2018gpem} -- a grammar-based NeuroEvolution approach. The CNN optimised by DENSER was evaluated using a fixed learning rate strategy, and thus it is likely that better learning policies exist. The architecture was generated for the CIFAR-10 dataset using a fixed learning rate of 0.01, where the individuals were trained for 10 epochs. The details of how the network was created are important as they might inform our conclusions later on. The specific topology of the network is described in Figure~\ref{fig:cnn_topology}. 

\subsection{Dataset}
\label{sec:dataset}
We opted to use the Fashion-MNIST instead of the network's native CIFAR-10 as it is a dataset where the training is faster. This dataset is composed by 70000 instances: 60000 for training and 10000 for testing. Each instance is $28 \times $28 grayscale image, which contrasts with CIFAR-10's $32 \times $32 RGB images. We will be scaling our images into $32 \times $32 RGB as they would not fit the network's input layer otherwise. This scaling was performed using the nearest neighbour method, and to pass from one to three channels we replicate the single channel three times.

\begin{figure}[t!]
    \centering 
    \includegraphics[width=0.35\textwidth,keepaspectratio]{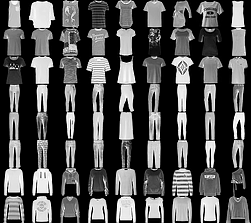}
    \caption{Example images from the Fashion-MNIST dataset.}
    \label{fig:fashionmnist} 
\end{figure}

\subsection{Experimental Setup}
\label{sec:experimental_setup}

We divide the experimental setup into two parts: the parameters used for the evolutionary search (Section~\ref{sec:evolution}); and for a longer training after the end of evolution (Section~\ref{sec:testing}).

\subsubsection{Evolution}
\label{sec:evolution}

\begin{table}[t!]
    \centering
    \begin{tabular}{c | c}
        \textbf{\gls{SGE} Parameter} & \textbf{Value} \\ \hline
        Number of runs & 10 \\ 
        Number of generations & 50 \\  
        Number of individuals & 5 \\
        Mutation rate &  0.15\\ \hline
        \textbf{Dataset Parameter} & \textbf{Value} \\ \hline
        Training set &  7000 instances from the training\\
        Validation set & 1500 instances from the training \\ 
        Test set &  1500 instances from the training \\ \hline
        \textbf{Training Data Augmentation} & \textbf{Value} \\ \hline
        Feature-wise Center & True \\
        Feature-wise Std. Deviation & True \\
        Rotation Range & 20 \\
        Width Shift Range & 0.2 \\
        Height Shift Range & 0.2 \\
        Horizontal Flip & True \\ \hline
        \textbf{Early Stop} & \textbf{Value} \\ \hline
        Patience & 3 \\
        Metric & Validation Loss \\
        Condition & Stop if Validation Loss does not\\
        & improve in 3 consecutive epochs \\ \hline
        \textbf{Network Training Parameter} & \textbf{Value} \\ \hline
        Batch Size & 1000 \\ 
        Epochs & 100 / 20 \\ 
        Metrics & Accuracy \\ 
    \end{tabular}
    \vspace{6pt}
    \caption{Experimental parameters.}
    \label{tab:exp_parameters}
\end{table}

The experimental parameters are summarised in Table~\ref{tab:exp_parameters}. They are organized into five sections:
\begin{itemize}
    \item \gls{SGE} Parameters -- parameters of the evolutionary engine.
    \item Dataset Parameters -- number of instances of each of the data partitions.
    \item Early Stop  -- the stop condition used to halt the training of the \gls{ANN}.
    \item Training Data Augmentation -- real time data augmentation parameters.
    \item Network Training Parameters -- parameters used when training the ANN.
\end{itemize} 

Our experimental parameters were picked with some considerations. Since evolutionary algorithms are very demanding in terms of computation resources it was paramount that the parameters used allowed us to perform meaningful evolutionary runs that could be completed in an acceptable time-frame. This motivated the selection of parameters that effectively reproduce an evolutionary strategy. Additionally, the fitness function operates on a fraction of the dataset as training utilizing all 60000 training examples was too time consuming. We also picked the training parameters accordingly.
Ideally, we would perform evolution on 100 training epochs with no early stop as we are trying to optimize the network's performance as much as possible. Instead, we performed two sets of experiments: (i) using 100 epochs and an early stop mechanism; (ii) using 20 epochs with no early stop. We started by reducing the computational cost through the implementation of an early stop mechanism. Notwithstanding, we were concerned that the evolutionary process would exploit this mechanism, which motivated the 20 epochs experience, where no early stop is used and the cost is instead reduced by reducing the training epochs.   
\subsubsection{Testing}
\label{sec:testing}

\begin{table}[t!]
    \centering
    \begin{tabular}{c | c}
        \textbf{Dataset Parameter} & \textbf{Value} \\ \hline
        Train set &  52500 instances from training data \\
        Validation set & 7500 instances from training data \\ 
        Test set &  10000 instances from test data \\ \hline
    \end{tabular}
    \vspace{6pt}
    \caption{Dataset information for parameters.}
    \label{tab:data_parameters}
\end{table}

After the evolutionary process is complete we need to properly assess the quality of the generated policies. 
The testing routine is the same as our fitness function, differing only in the data used (seen in Table \ref{tab:data_parameters}). 

In our testing routine we use all training instances (splitting them into training and validation) to train the network using the policy we want to evaluate. This network is subsequently tested using all test data (that was not used previously) to obtain an unbiased \textbf{test accuracy}. We will also be tracking each policy's best \textbf{validation accuracy} to have additional insight into how well the learned weights are able to generalize. Finally, we need to decide on a policy to serve as a baseline. We chose to use a \textit{static learning rate policy of 0.01} for three reasons. The fact that the network was evolved using this learning rate (as was explained in Section~\ref{sec:network}). This assures us that this is an adequate learning rate for this network making it a good benchmark for our evolved policies. Additionally, this particular constant is the most common policy in often used Deep Learning frameworks~\cite{keras, matlab}. We believe that benchmarking against such a widely used policy is a proper way to test our hypothesis. Finally, we had to use a baseline that had similar information to our evolved methods. The adaptive techniques referred to in Section~\ref{sec:lr_optimization}, for example, use the gradient of the loss function to make more precise adjustments to a per-dimension learning rate. The fact that these methods have access to additional information means they are not suitable as benchmarks. 

We have three testing scenarios:
\begin{itemize}
    \item The \textbf{first scenario (1)} is the same as the first evolutionary scenario, i.e., training is done for \textbf{100 epochs with the early stop mechanism}. 
    \item The \textbf{second scenario (2)} trains for only \textbf{20 epochs, with no early stop}.
    \item The \textbf{third scenario (3)} trains for \textbf{100 epochs, but the early stop mechanism is disabled}.
\end{itemize}

The first and second scenarios exist primarily so we can see how the evolved policies compare with the baseline in the conditions they were evolved in. Scenario 3 yields the most important results as its conditions represent the typical use case of a neural network.

In order to make discussion clearer the evolved polices will be referred to as policy A (for the best policy evolved with the early stop mechanism) and policy B (for the best policy evolved with no early stop). These evolved polices were tested in their evolutionary environments (scenario 1 and 2 for policies A and B respectively) and in scenario 3. The baseline policy was tested in all 3 scenarios.

\subsection{Experimental Results}
\label{sec:experimental_results}

\begin{table}[]
\centering
\begin{tabular}{|c|c|c|c|c|}
\hline
\multicolumn{2}{|c|}{}                           & \multicolumn{3}{c|}{Policy}                                                                       \\ \cline{3-5} 
\multicolumn{2}{|c|}{\multirow{-2}{*}{Scenario}} & A                                               & B                      & Baseline               \\ \hline
                        & Validation         & $0.751 \pm 0.167$                                  & n/a          & \boldmath$0.859 \pm 0.003$\unboldmath \\ \cline{2-5}    
\multirow{-2}{*}{1}         & Test               & $0.692 \pm 0.241$                                   & n/a          & \boldmath$0.850 \pm 0.004$\unboldmath \\ \hline
                            & Validation         & n/a & $0.854 \pm 0.009$          & \boldmath$0.856 \pm 0.004$\unboldmath\\ \cline{2-5} 
\multirow{-2}{*}{2}         & Test               & n/a                     
& \boldmath$0.848 \pm 0.007$\unboldmath & $0.844 \pm 0.002$          \\ \hline
                            & Validation         & \boldmath$0.894 \pm 0.004$\unboldmath                          & $0.891 \pm 0.003$          & $0.888 \pm 0.002$          \\ \cline{2-5} 
\multirow{-2}{*}{3}         & Test               & \boldmath$0.887 \pm 0.002$\unboldmath                         & $0.854 \pm 0.009$          & $0.875 \pm 0.004$         \\ \hline

\end{tabular}
    \caption{Accuracy of the evolved policies (A \& B) on their evolutionary environment (1 \& 2 respectively) and scenario 3 (representative of an actual use case), compared with the baseline policy.}
    \label{tab:test_results}
\end{table}

The table presented in \ref{tab:test_results} summarizes the results of our experimentation, showing the average and standard deviation of the accuracy of a given policy in a specific scenario over five runs. As detailed in Section~\ref{sec:testing}, each run trains the network using the chosen policy and subsequently tests its accuracy on the 10000 test instances.

\subsubsection{Scenario 1} yields results that are not intuitive given the circumstances. Training in this scenario can be halted by an early stop mechanism. Since policy A was evolved using this same kind of training it is to be expected that it would perform well in these conditions. However, the results show the opposite. Policy A, in fact, performs far worse then the baseline when early stop is in use. Analysing individual results showed that this policy will occasionally trigger the early stop in the first few epochs (this can be observed in the large standard deviation associate with these trials). There are several interpretations for the implications this has on the validity of the evolutionary process. On the one hand it can be argued that this demonstrates an issue with the evolutionary process since the policy is not a consistent solution to the problem it is supposed to solve. While it is a fact that the policy is an inconsistent solution we do not believe this implies any problems with the evolution. The fact that this policy can, on occasion, yield the best performance implies the genetic information of this individual is useful for the evolutionary process. 

\subsubsection{Scenario 2} results are more in line with our expectations. We can observe that, albeit only marginally, policy B shows better test accuracy than the baseline when trained under the parameters it was evolved for. It is noteworthy that policy B does not have superior accuracy in validation. This suggests that the \textbf{evolved policy is outperforming the baseline in its ability to generalize} when moved to a different set of data. 

\subsubsection{Scenario 3} was designed to test which policy is able to get the most out of this network's architecture and it gave the most important set of results. The results show that, under these conditions, \textbf{the best accuracy this network achieved was obtained using an evolved policy for training}. On average, policy A performs better than the baseline in the test set by 1.2\% and it obtains these good results more consistently. Another interesting result is that policy B (that was previously outstanding because of its ability to generalize) suffers the biggest dip in performance from validation to test in this scenario. Ideally, both evolved policies would outperform the baseline. There are, however, some possibly limiting factors. Namely, it is possible that the shorter training duration used in scenario 2 discourages the evolution of policies that translate well into scenario 3. This topic is discussed further in \ref{sec:shape} as we analyse policy B's shape. 

\subsubsection{Shape}
\label{sec:shape}
\begin{figure}[t!]
    \centering 
    \includegraphics[width=0.5\textwidth,keepaspectratio]{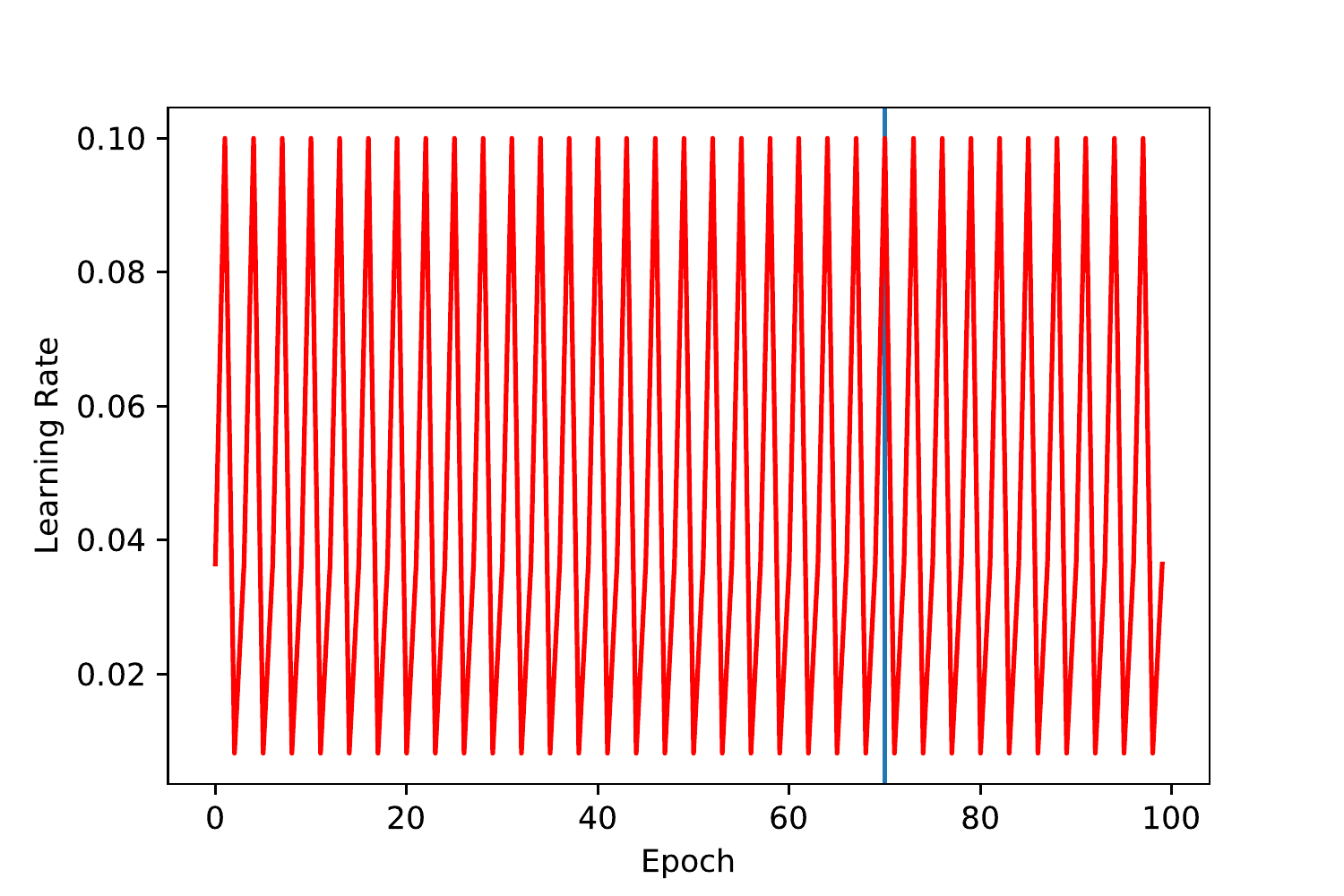}
    \caption{Policy A}
    \label{fig:policyA} 
\end{figure}

\begin{figure}[t!]
    \centering 
    \includegraphics[width=0.5\textwidth,keepaspectratio]{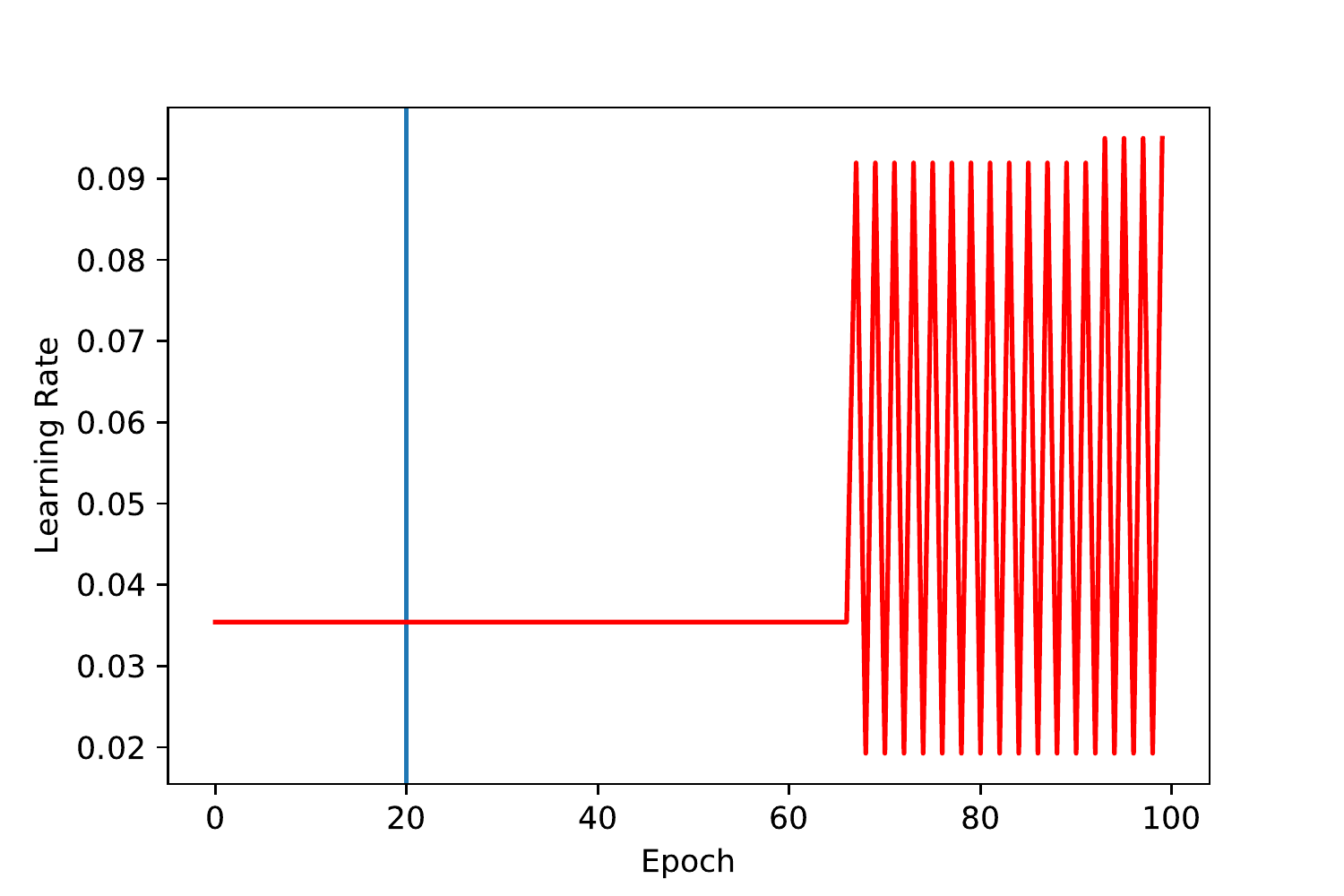}
    \caption{Policy B}
    \label{fig:policyB} 
\end{figure}

As discussed in \ref{sec:introduction}, we are interested in analysing the shapes that our evolved policies take. Namely, in this section we will be analysing the shape of the previously discussed Policies A and B. These policies can be observed in Figures \ref{fig:policyA} and \ref{fig:policyB}. These figures show how the learning rate evolved over time as well as a vertical line that signals the epoch where the training using this policy stopped.

Observing the shape of policy A (seen in Figure \ref{fig:policyA}) led to some interesting insights. Initially, it seemed that this policy only had the best performance during evolution because its shape cheat the early stop mechanism. We suspected that by frequently using a high learning rate it might be possible to create false improvements that trick the system. To elaborate, it is feasible for a policy to routinely worsen and subsequently improve its performance \textit{on purpose} in order to pass the early stop check.  We can, in fact, observe that this policy is able to train for a long time despite the early stop as the vertical line shows. As a comparison, the baseline policy typically triggers the early stop between epochs 20-30, which means that policy A is able to train for twice as long. 

Policy B took on a very different shape. Despite the erratic behaviour shown past epoch 60, this policy is effectively a static learning rate as its training always ended in epoch 20 (as a reminder, no early stop was used in the evolution of the policy). While this initially seems disappointing (finding an adequate constant is not something that requires such a complex system), it is important to understand that, due to the reduced training duration, there is a possibility that the benefits of using a dynamic policy in this context are negligible, stifling probability that they show up in evolution. The idea that the evolution of dynamic methods is suppressed under these circumstances is further supported by the fact that all twelve of the best policies during the evolution of policy B were constants. In this context the twelve best policies we are referring to is the set of policies that were, at some point during evolution, the best policy in all runs.

We have, up until this point, observed two types of evolved functions shapes (within the individuals that perform well). The first type is constants, these comprise the majority of the search space so their presence is expected. The second type can be observed in policy A, we refer to these as \textit{oscillator policies}. We believe that these policies are approximations of the policies used in \cite{smith2017cyclical}. While it would be disingenuous to claim that we are evolving cyclical policies, it seems feasible that the evolved oscillator policies are effective for the same reasons as the cyclical ones. It is notable that while many known dynamic policies are decaying policies we have not observed any well performing evolved policies with a similar shape.

\section{Conclusions and Future Work}
\label{sec:conclusions}

In this work we posed the question of whether or not evolving learning rate policies was a viable way of improving a network architecture's performance. To test this, we designed and developed AutoLR. a framework that optimizes learning rate policies using \gls{SGE}. Furthermore, this framework was then utilized to create two evolved policies. These evolved policies were tested and compared with a widely used baseline policy. Both of the policies evolved were able to improve on the established baseline in some capacity. Not only that, the network's best recorded performance was achieved with an evolved policy, suggesting that \textbf{evolving learning rate policies for a specific architecture did in fact improve the network performance}. Additionally, some of the evolved policies resemble man-made policies seen in \cite{smith2017cyclical}, suggesting that the system might have implicitly discovered the ideas that make such policies effective. In the future we would like to expand the range of policies that can be evolved to enable meaningful comparisons with a wider array of state of the art methods.

\section{Acknowledgements}
This work is funded by national funds through the FCT - Foundation for Science and Technology, I.P., within the scope of the project CISUC - UID/CEC/00326/2020 and by European Social Fund, through the Regional Operational Program Centro 2020. \\
Filipe Assunção is partially funded by: Fundação para a Ciência e Tecnologia (FCT), Portugal, under the PhD grant\\ SFRH/BD/114865/2016. We also thank the NVIDIA Corporation for the hardware granted to this research.

\bibliographystyle{ACM-Reference-Format}
\bibliography{sample-bibliography} 

\end{document}